\documentclass[10pt,twocolumn,letterpaper]{article}

\usepackage{cvpr}              

\usepackage{multirow}
\usepackage{tabularx}
\usepackage{booktabs}
\usepackage[most]{tcolorbox}
\usepackage{enumitem}
\usepackage{color}
\usepackage{tabularray}
\usepackage{longtable} 
\usepackage{array}     

\NewDocumentCommand{\todo}
{ mO{} }{\textcolor{magenta}{\textsuperscript{\textit{TODO}}\textsf{\textbf{\small[#1]}}}}

\definecolor{cvprblue}{rgb}{0.21,0.49,0.74}
\usepackage[pagebackref,breaklinks,colorlinks,allcolors=cvprblue]{hyperref}

\def\paperID{***} 
\def\confName{CVPR}
\def\confYear{2026}

\title{Benchmarking PhD-Level Coding in 3D Geometric Computer Vision}

\author{
Wenyi Li\textsuperscript{1*}, 
Renkai Luo\textsuperscript{1,2*}, 
Yue Yu\textsuperscript{1,3,6}, 
Huan-ang Gao\textsuperscript{1}, 
Mingju Gao\textsuperscript{1}, \\ 
Li Yuan\textsuperscript{4}, 
Chaoyou Fu\textsuperscript{5}, 
Hao Zhao\textsuperscript{1,3$\dagger$} \\
\textsuperscript{1}AIR, Tsinghua University ~
\textsuperscript{2}Qiuzhen College, Tsinghua University~ \\
\textsuperscript{3}BAAI ~
\textsuperscript{4}Peking University ~
\textsuperscript{5}Nanjing University ~
\textsuperscript{6}University of Toronto\\
{\tt\small  \url{https://geocodebench.github.io/}}
}

\begin{document}
\maketitle
\begin{abstract}

\renewcommand\thefootnote{}
\footnotetext{\textsuperscript{*}Equal Contribution,\textsuperscript{$\dagger$}Corresponding Author} 

AI-assisted coding has rapidly reshaped software practice and research workflows, yet today’s models still struggle to produce correct code for complex 3D geometric vision. If models could reliably write such code, the research of our community would change substantially. To measure progress toward that goal, we introduce GeoCodeBench, a PhD-level benchmark that evaluates coding for 3D vision. Each problem is a fill-in-the-function implementation task curated from representative papers at recent venues: we first let a tool propose candidate functions from official repositories, then perform careful human screening to select core 3D geometric components. For every target, we generate diverse, edge-case unit tests, enabling fully automatic, reproducible scoring. We evaluate eight representative open- and closed-source models to reflect the current ecosystem. The best model, GPT-5, attains only 36.6\% pass rate, revealing a large gap between current capabilities and dependable 3D scientific coding. GeoCodeBench organizes tasks into a two-level hierarchy: General 3D capability (geometric transformations and mechanics/optics formulation) and Research capability (novel algorithm implementation and geometric logic routing). Scores are positively correlated across these axes, but research-oriented tasks are markedly harder. Context ablations further show that “more paper text” is not always better: cutting off at the Method section statistically outperforms full-paper inputs, highlighting unresolved challenges in long-context scientific comprehension. Together, these findings position GeoCodeBench as a rigorous testbed for advancing from generic coding to trustworthy 3D geometric vision coding.

\end{abstract}    
\section{Introduction}
\label{sec:intro}

Coding has emerged as one of the most mature and reliable application areas of modern large language models (LLMs)~\cite{chen2021evaluating,austin2021program,jimenez2023swe}. 
Yet for our community, current LLMs remain far from being able to write correct, complete, and trustworthy code for complex 3D geometric vision. In contrast to generic software engineering, scientific 3D vision demands mathematically precise geometric operators~\cite{liu2025hogs, gao2025curve, guo2024tetsphere}, physically grounded formulations~\cite{ye2025geosplatting, gu2025irgs, yao2024reflective}, and reliable reasoning about multi-view, multi-modal data~\cite{zhou2025mgsr, liu2025monosplat}. If LLMs were able to generate such code reliably, the impact would be transformative: automating model prototyping, accelerating research cycles, and democratizing 3D algorithm development.

\textbf{GeoCodeBench} is the first step toward that long-term vision, which is a benchmark designed to answer a foundational question: How well can LLMs implement real 3D vision code when given the same textual context that human researchers use? Our benchmark follows the end-use scenario directly. Each problem presents a function skeleton extracted from a real research pipeline (e.g., geometric component, rendering logic, physics-based formulation, or novel algorithm module) together with the relevant text from a paper (analogous to how future users might describe a research idea). The model must fill in the missing implementation, and correctness is judged purely by executing the code in a sandbox with unit tests. This setup mirrors how an LLM would be deployed in practice: reading scientific text, interpreting mathematical definitions, implementing nontrivial logic, and passing rigorous tests.

Constructing such a benchmark is challenging because 3D vision implementations require deep domain expertise, and no existing dataset meets this requirement. GeoCodeBench fills this gap as the first PhD-level coding benchmark for 3D geometric vision. To minimize data leakage, we curate problems exclusively from recent top-venue 3D vision papers. We begin by prompting an automated toolchain to propose candidate functions from official repositories, but we find that many candidates are trivial, auxiliary, or unsuitable for evaluation. Therefore, we introduce a meticulous \textbf{expert-in-the-loop} process: 3D vision researchers manually inspect the repositories, apply strict selection criteria, and identify core geometric or algorithmic functions suitable for evaluation. The result is a collection of high-fidelity targets that reflect the actual building blocks of contemporary 3D vision systems.

A critical component of coding benchmarks is the unit-test infrastructure. The quality of the evaluation depends not only on the task itself but on the diversity, extremity, and coverage of its tests. For GeoCodeBench, we provide customized, high-variability unit tests for every problem: probing edge cases in geometry, conditioning on degenerate configurations, and validating invariances.

Our design philosophy organizes tasks into two major categories and four fine-grained capabilities, as highlighted in Fig.~\ref{fig:teaser}. (1) General 3D Capability: (1.1) Geometric transformations: coordinate conversions, projections, normals, and other operations foundational to all 3D pipelines. (1.2) Mechanics/optics formulation: implementing analytic optics, mechanics equations, or radiometric operators. (2) Research Capability: (2.1) Novel algorithm implementation: functions that realize a paper’s new idea. (2.2) Geometric logic routing: creative recombination of existing operators, reflecting how many influential papers construct new pipelines by composing classic building blocks. This taxonomy captures both the fundamentals of 3D geometry and the higher-level reasoning needed for research-grade code.

We evaluate eight representative open and closed source models (Fig.~\ref{fig:teaser}) across the entire benchmark. Several phenomena emerge. \textbf{First}, the best model, GPT-5, reaches only 36.6\% pass rate, revealing a substantial gap between current capabilities and reliable 3D scientific coding. \textbf{Second}, successful cases often display creative correctness: models implement an algorithm using a completely different but mathematically valid approach (see Fig.~\ref{fig:teaser} top-right) that still passes all unit tests, which shows potential for genuine problem-solving beyond reproduction of ground-truth code. \textbf{Third}, Research tasks are more difficult than General tasks, yet the two axes are positively correlated, suggesting that strong geometric fundamentals are necessary but not sufficient for research-level implementation. \textbf{Fourth}, providing more context is not always beneficial: truncating papers at the Method section outperforms full-paper input, highlighting challenges in long-context scientific comprehension.
\section{Related Work}
\label{sec:relatedwork}

\providecommand{\gcheck}{\ensuremath{\surd}}
\providecommand{\ycheck}{\ensuremath{\surd}}
\providecommand{\rxmark}{\ensuremath{\times}}
\providecommand{\toprule}{\hline}
\providecommand{\midrule}{\hline}
\providecommand{\bottomrule}{\hline}
\providecommand{\todo}[1]{[TODO: #1]}

\makeatletter
\@ifundefined{textcolor}{}{%
  \renewcommand{\gcheck}{\textcolor{green}{\ensuremath{\surd}}}%
  \renewcommand{\ycheck}{\textcolor{yellow}{\ensuremath{\surd}}}%
  \renewcommand{\rxmark}{\textcolor{red}{\ensuremath{\times}}}%
}
\makeatother

\paragraph{Multi-modal Foundation Model Benchmarks.}
As multimodal large language models (MLLMs) advance, systematic evaluation across task families is crucial for diagnosing capability and generalization; large-scope efforts emphasize coverage and dataset design principles \cite{yue2024mmmu, li2024seed, li2025information}.
VRBench, LVBench, and MLVU \cite{yu2025vrbench, wang2025lvbench, zhou2025mlvu} probe long-horizon video understanding and temporal reasoning.
MMReason, ReasonVQA, and ProJudge \cite{yao2025mmreason, tran2025reasonvqa, ai2025projudge} assess open-ended, multi-step reasoning, with ProJudge in particular targeting process-level judging; structured variants test stepwise CoT with visual programs, interleaved multi-image, and cross-source composition \cite{gao2025benchmarking, du2025easy, tian2025mmcr}. 
Beyond single turns, \emph{MultiVerse} benchmarks multi-turn conversations with vision-language models~\cite{lee2025multiverse}, whereas \emph{m\&m's} and \emph{OmniACT} focus on tool-augmented agents that plan and execute multi-step computer or API-based actions from multimodal inputs~\cite{kapoor2024omniact}, while
safety and robustness are scrutinized via safety pitfalls, hallucination, and adversarial stress tests \cite{tu2023many, kaul2024throne, zhao2025one}.
Domain coverage is also broadened, spanning from medical imaging and clinical reasoning \cite{liu2025gemex, hu2024omnimedvqa, sun2024pathmmu} to industry settings such as insurance, agriculture, and driving \cite{lin2025ins, shinoda2025agrobench, park2025nuplanqa}.
Despite this breadth, most settings remain query-oriented with targeted inputs and little long-context, cross-source composition, prioritizing text answers over executable, code-grounded solutions, which motivates a benchmark that couples long-context multi-step reasoning with coding proficiency.

\paragraph{Code Generation Benchmarks.}
Existing code generation benchmarks increasingly emphasize execution-based evaluation to measure the true programming capability of Large Language Models (LLMs). Early datasets such as HumanEval and MBPP established the docstring-to-function synthesis paradigm and pass@k evaluation \cite{chen2021evaluating,austin2021program}, later extended by APPS for competition-level reasoning \cite{hendrycks2021measuring}, MultiPL-E for multi-language robustness \cite{cassano2022multipl}, and SWE-bench for repository-level issue resolution under real dependencies \cite{jimenez2023swe}. More recently, LiveCodeBench addressed contamination and saturation \cite{jain2024livecodebench}, while DomainCodeBench revealed limited generalization of top models to specialized domains \cite{zheng2024top}. Parallel to these general-purpose efforts, research-agent benchmarks shift toward end-to-end scientific reproduction. PaperBench evaluates from-scratch replication of ICML papers with structured rubrics \cite{starace2025paperbench}; ResearchCodeBench masks contribution-critical code fragments and scores execution-grounded replication \cite{hua2025researchcodebench}; and CORE-Bench \cite{siegel2024core}, SUPER \cite{bogin2024super}, RECODE-H \cite{miao2025recode}, SciReplicate-Bench \cite{xiang2025scireplicate}, and Paper2Code \cite{seo2025paper2code} explore reproducibility, repository setup, multi-turn feedback, LaTeX-to-code generation, and paper-to-repository synthesis. Despite this breadth, existing benchmarks are predominantly contest-oriented or focus on holistic paper reproduction, with limited coverage of domain-specific scientific implementation. None simultaneously targets 3D geometric reasoning, long-context paper-to-code mapping, multi-step executable synthesis, and hidden-test robustness—gaps that \textbf{GeoCodeBench} is designed to fill.

\paragraph{GeoCodeBench Questions.}

Building on these trends, GeoCodeBench grounds its geometric track in recent 3D vision literature, using representative papers to define tasks, baselines, and evaluation references.
Inverse rendering: inter-reflection ray-traced GS \cite{gu2025irgs}, geometry-/PBR-guided splatting \cite{ye2025geosplatting,RefGaussian}, relightable SDF \cite{DiscretizedSDF}, 2D/3D mutual boosting \cite{zhou2025mgsr}, illumination-agnostic NVS \cite{zhou2025lita}. Camera/optics: distorted cameras, motion blur, large FOV, unbounded and controllable DoF via \cite{wu20253dgut,lee2025comogaussian,deng2025self,liu2025hogs,shen2025dof,wang2025dof,song2024sa,yuan2024slimmerf}. Panorama: omni/panorama feed-forward GS \cite{lee2025omnisplat,chen2025splatter,zhang2025pansplat}. Sparse/generalization/priors: self-ensembling, registration, geometry/epipolar densification, UV/mono-depth/diffusion priors, learning-free semantic lifting \cite{zhao2025self,cheng2025reggs,wu2025sparse2dgs,zheng2025nexusgs,rai2025uvgs,liu2025monosplat,zhong2025taming,marrie2025ludvig}. Dynamics/SLAM/sensors: 4DGS SLAM, event/LiDAR pipelines, dynamic-static decomposition, explicit motion embeddings \cite{li20254d,huang2025inceventgs,giacomini2025splat,wang2025degauss,wei2025emd,liu2024rip}. Efficiency/compression: sparsifying, localized management, fast schedules, elastic inference, feed-forward compression, 2D-GS, hardware-rasterized RayGS \cite{zhang2025gaussianspa,yang2025improving,chen2025dashgaussian,liu2025flexgs,chen2024fast,zeng2025instant,bulo2025hardware}. Geometry/editing/apps: curves, compositional/editable hybrids, TetSphere meshes, riggable heads, polarimetry, BEV perception \cite{gao2025curve,jiang2024gaussianblock,guo2024tetsphere,lee2024surfhead,han2025polgs,lu2025toward,wu2023mars}.
GeoCodeBench distills these works into unit-testable geometric and rendering operators that demand 3D reasoning and precise implementation, ensuring evaluation reflects both understanding and coding proficiency.

\begin{table}[t]
  \centering
  \scriptsize             
  \setlength{\tabcolsep}{2.5pt}  
  \renewcommand{\arraystretch}{1.15}
  \caption{\textbf{Capability Coverage across Representative Benchmarks and GeoCodeBench.}(~\gcheck~strong/explicit; \ycheck~partial/incidental; \rxmark~not a primary focus.)}
  \label{tab:capability_coverage}
  \begin{tabular}{@{}lccccccc@{}}
    \toprule
    {\textbf{Capability}} &
    {\tiny \textbf{MMMU~\cite{yue2024mmmu}}} &
    {\tiny \textbf{VR-B~\cite{yu2025vrbench}}} &
    {\tiny \textbf{MBPP~\cite{austin2021program}}} &
    {\tiny \textbf{SWE~\cite{jimenez2023swe}}} &
    {\tiny \textbf{Paper-B~\cite{starace2025paperbench}}} &
    {\tiny \textbf{RCB~\cite{hua2025researchcodebench}}} &
    {\tiny \textbf{Ours}} \\
    \midrule
    Func. synthesis       & \rxmark & \gcheck & \gcheck & \gcheck & \rxmark & \gcheck & \gcheck \\
    Paper-to-code      & \rxmark & \rxmark & \rxmark & \rxmark & \gcheck & \gcheck & \gcheck \\
    Format discipline & \rxmark & \rxmark & \ycheck & \gcheck & \gcheck & \gcheck & \gcheck \\
    Domain knowledge     & \gcheck & \rxmark & \rxmark & \gcheck & \gcheck & \gcheck & \gcheck \\
    Hidden tests & \rxmark & \rxmark & \rxmark & \gcheck & \ycheck & \gcheck & \gcheck \\
    3D impl.   & \rxmark & \rxmark & \rxmark & \rxmark & \rxmark & \rxmark & \gcheck \\
    \bottomrule
  \end{tabular}

\end{table}

\begin{figure*}[t]
    \centering
    \includegraphics[width=0.95\linewidth]{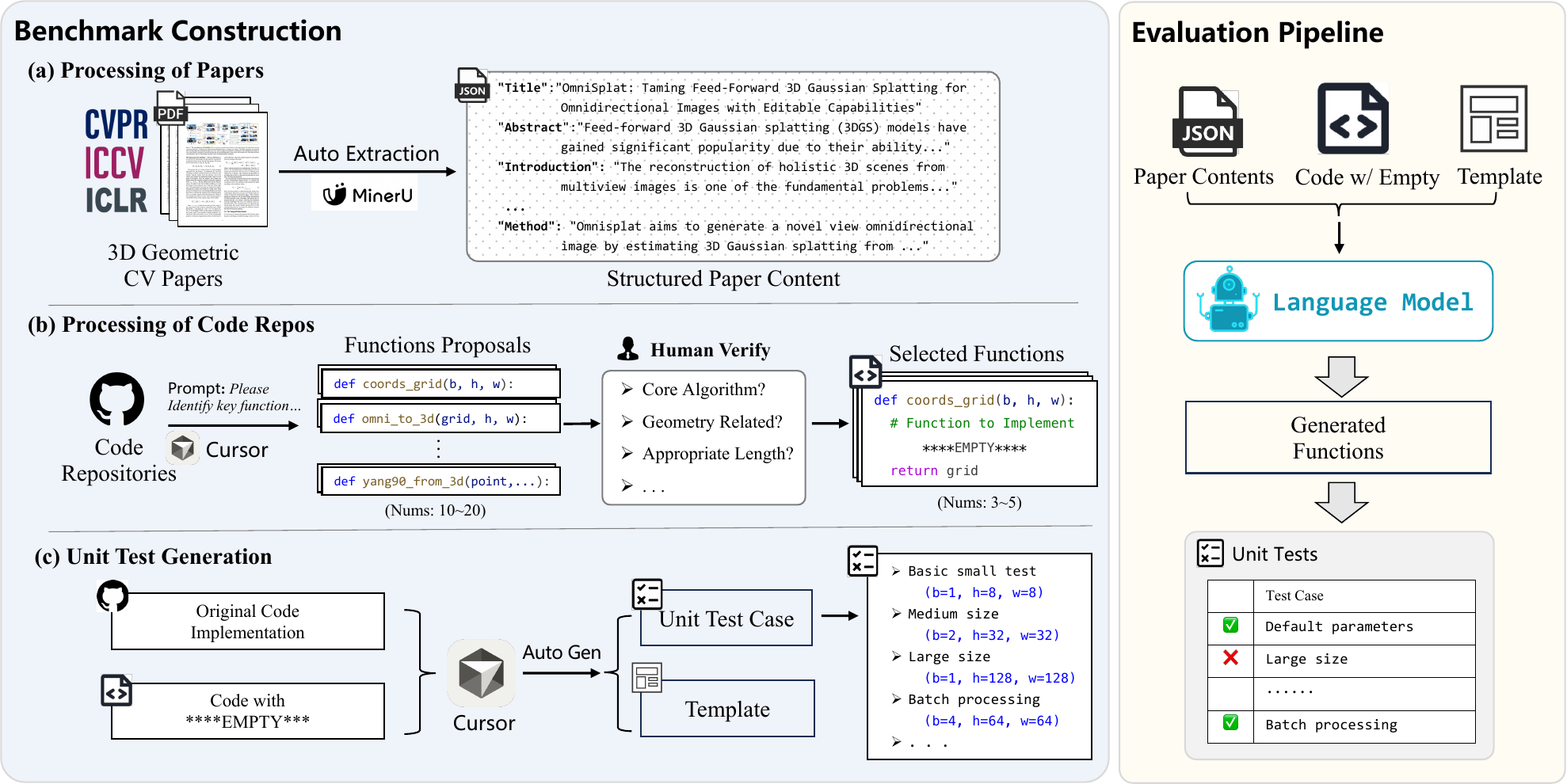}
    \caption{The Benchmark Curation and Evaluation Pipeline of GeoCodeBench. }
    \label{fig:pipeline}
\end{figure*}

\section{GeoCodeBench}
\label{sec:method}
\subsection{Overview}
\textbf{GeoCodeBench} is a benchmark designed to evaluate whether large language models can understand and implement algorithms from recent 3D vision research through executable code. 
It consists of 100 high-quality fill-in-the-blank code completion problems, curated from the latest top-tier papers and their corresponding open-source repositories.
Each problem includes structured paper content, a partially masked source-code function, and corresponding unit tests. 
GeoCodeBench features realistic research tasks, code-level verification, and a hierarchical evaluation protocol covering both general 3D understanding and research-oriented implementation.

\subsection{Benchmark Construction}

GeoCodeBench focuses on a wide range of 3D vision domains, including Gaussian Splatting, Pose Estimation, SLAM, Reconstruction,  Physics-based Modeling, NeRF, and 3D Segmentation.
We collect benchmark problems from recent papers published at top-tier venues such as CVPR, ICCV, and ICLR, together with their corresponding open-source repositories. To reduce the risk of instances appearing in LLM pretraining data, we select only papers published in 2025. As illustrated in Figure~\ref{fig:pipeline}, the construction process begins with collecting and preprocessing both research papers and their associated code repositories. 




\paragraph{Automatic Processing of Paper PDFs.} 
Since LLMs cannot directly parse raw PDF, for each paper, we use a state-of-the-art OCR model, MinerU~\cite{mineru}, to automatically extract text, equations, and figures from the PDF. 
We convert the extracted content into a structured JSON format organized by sections( \texttt{\{"Title":"...", "Introduction":"...", "Method":"..." \}}), and separately store all figures for reference. 

\paragraph{Automatic Processing of Code Repositories.} 
For the corresponding code repositories, we prompt Cursor~\cite{cursor} to automatically suggest candidate functions that potentially implement key algorithmic components, typically producing 10–20 candidates per repository. 
Human experts then review and refine these automatically proposed functions based on their relevance to the paper’s core algorithm and the feasibility of the subsequent evaluation. 
After this selection, around 3–5 high-quality functions are retained per repository, and their implementation bodies are masked using the placeholder \texttt{****EMPTY****} to form executable code-completion tasks.


\paragraph{Unit Test Generation.}
For fill-in-the-blank code completion problems, the most reliable way to verify the correctness of generated implementations is through unit testing. 
In practice, since the original repositories rarely provide function-scoped tests for the selected targets, we have to construct function-level tests for each target within GeoCodeBench.
However, designing comprehensive unit tests manually is time-consuming and labor-intensive. To reduce manual effort and improve scalability, we employ automated tools to streamline the process. 
Given the original implementation and its masked version (with the target function body replaced by \texttt{EMPTY}), we prompt Cursor~\cite{cursor} to automatically generate 10 test cases that evaluate correctness under diverse parameter configurations, as illustrated in Figure~\ref{fig:pipeline}.
Alongside the test cases, Cursor also produces a standardized template that includes the required imports, input–output definitions, and other necessary elements. This template constrains the expected answer format from LLMs and facilitates automatic extraction and execution during evaluation.
Each unit test supplies identical inputs to both the original function and the model-generated version, and asserts that their outputs match. 
The tests cover both default and edge conditions to assess functional fidelity and generalization. All auto-generated unit tests are subsequently reviewed by human experts to guarantee reliability.

\subsection{Evaluation Pipeline}
As shown in Figure~\ref{fig:pipeline}, each input to the LLM in GeoCodeBench comprises three key components:
(1) \textbf{Structured Paper Content:} the entire paper is parsed and organized into a structured JSON format.
(2) \textbf{Code with Masked Function:} the source file containing the target function is provided as input, where the target function body is replaced with the placeholder \texttt{****EMPTY****}.
(3) \textbf{Unified Execution Template:} a standardized template defines input–output interfaces, imports, and necessary elements for unit testing.
These inputs are fed into the target LLM, which is required to generate the missing implementation.
The produced code is then executed against the benchmark’s unit tests, providing a fully automatic, reproducible measure of performance.
We evaluate model performance using a \textit{pass rate} metric. 
For each instance $i$, let $p_i$ and $T_i$ denote the number of passed and total test cases, respectively. 
The overall benchmark score is defined as:
\begin{equation}
    \text{PassRate} = \frac{1}{N} \sum_{i=1}^{N} \frac{p_i}{T_i}.
\end{equation}



\subsection{Benchmark Statistics}

\paragraph{Taxonomy.}
To better characterize the capability spectrum that \textbf{GeoCodeBench} aims to evaluate, we define a two-level taxonomy reflecting the essential competencies required for understanding and implementing 3D vision algorithms. 
\textbf{(1) General 3D Capability} focuses on geometric reasoning skills, including \emph{Geometric Transformations} (24\%) and \emph{Mechanics/optics} (31\%), which cover coordinate conversions, rotation parameterizations, projection models, and differentiable equations. 
\textbf{(2) Research Capability} evaluates higher-level reasoning and adaptation abilities, including \emph{Novel Algorithm Implementation} (34\%) and \emph{Geometric Logic Routing} (11\%), which translate mathematical principles into executable procedures and combine multiple modules into complete systems. 
Detailed statistics of GeoCodeBench are provided in the supplementary material.


\subsection{Features of GeoCodeBench}

\paragraph{Challenging and Real-world Tasks.}
GeoCodeBench derives tasks directly from real research papers and official repositories. 
Each problem represents an authentic algorithmic component in modern 3D vision studies, providing a realistic and challenging setting that reflects actual research workflows.

\paragraph{Automated and Scalable Pipeline.}
GeoCodeBench is constructed through an automated pipeline that parses papers, extracts functions, and generates unit tests with minimal human effort. 
The unified test framework supports large-scale extension to new research papers.

\paragraph{Community-driven Evolution.}
GeoCodeBench evolves continuously with the 3D vision community. 
Newly published papers can be incorporated through the same automatic pipeline, allowing the benchmark to grow alongside emerging research trends. 
This community-driven and extensible design establishes GeoCodeBench as a long-term foundation toward building automated research agents and, ultimately, an \emph{auto 3D vision scientist}.
\section{Diagnosing LLMs on GeoCodeBench}
\label{sec:experiment}


\begin{table*}[t]
\centering 
\small
\caption{
\textbf{The Performance of Representative LLMs on GeoCodeBench.} 
Performance is assessed across two aspects: \textit{General 3D Capability} (General) and \textit{Research Capability} (Research), four dimensions:  \textit{Geometric Transformations} (Geo. Trans.),  \textit{Mechanics/Optics Formulation} (Mech./Opt.), \textit{Novel Algorithm Implementation}(Algorithm), and \textit{Geometric Logic Routing}(Routing).}
\label{tab:main}
\renewcommand{\arraystretch}{1.05} 
\begin{tabularx}{\textwidth}{
    l|c|c|cc| cccc
}
\toprule
 & {} 
 & {} 
 & {} 
 & 
 & \multicolumn{2}{c}{\textbf{General 3D Capability}} 
 & \multicolumn{2}{c}{\textbf{Research Capability}} \\
\multirow{-2}{*}{\textbf{Model}} 
 & \multirow{-2}{*}{{\textbf{Company}}} 
 & \multirow{-2}{*}{\textbf{Overall}} 
 & \multirow{-2}{*}{\textbf{General}} 
 & \multirow{-2}{*}{\textbf{Research}} 
 & \textbf{Geo. Trans.} 
 & \textbf{Mech./Opt.} 
 & \textbf{Algorithm} 
 & \textbf{Routing} \\ \midrule


GPT-5 & OpenAI & \textbf{36.6\%} & \textbf{42.8\%} & \textbf{29.1\%} & 41.7\% & \textbf{43.7\%} & \textbf{29.1}\% & 28.9\% \\
Claude-Sonnet-4.5 & Anthropic  & 31.1\% & 37.2\% & 23.7\% & 38.3\% & 36.5\% & 19.7\% & \textbf{35.9\%} \\
Gemini-2.5-Pro & Google & 30.4\% & 33.8\% & 26.2\% & \textbf{41.9\%} & 27.6\% & 25.3\% & 29.1\%\\
Kimi-K2-Instruct & Moonshot & 30.4\% & 34.6\% & 25.1\% & 36.7\% & 33.1\% & 23.1\% & 31.4\% \\
Doubao-Seed-1.6 & ByteDance  & 26.9\% & 29.7\% & 23.4\% & 40.9\% & 21.0\% & 22.9\% & 25.2\% \\
Qwen3-Coder-480B & Alibaba  & 23.5\%& 22.7\% & 24.6\% & 29.0\% & 17.7\% & 21.8\% & 33.2\% \\
DeepSeek-R1 & DeepSeek & 21.0\% & 27.2\% & 13.5\% & 33.9\% & 21.9\% & 12.4\% & 17.0\% \\
Llama-3.1-405B-Instruct & Meta  & 14.3\% & 16.8\% & 11.3\% & 21.3\% & 13.2\% & 10.9\% & 12.7\% \\ \bottomrule

\end{tabularx}
\end{table*}

\subsection{Overall Model Comparison}
\label{subsec:overall}

We evaluate eight representative large language models (LLMs) on GeoCodeBench, encompassing both commercial and open-source systems.
Each model is required to generate directly executable 3D vision code based on given task instructions.
Performance is evaluated by the unit test pass rate across four dimensions, which together capture both general geometric understanding and research-level implementation capability.

\begin{figure}[h]
    \centering
    \includegraphics[width=\linewidth]{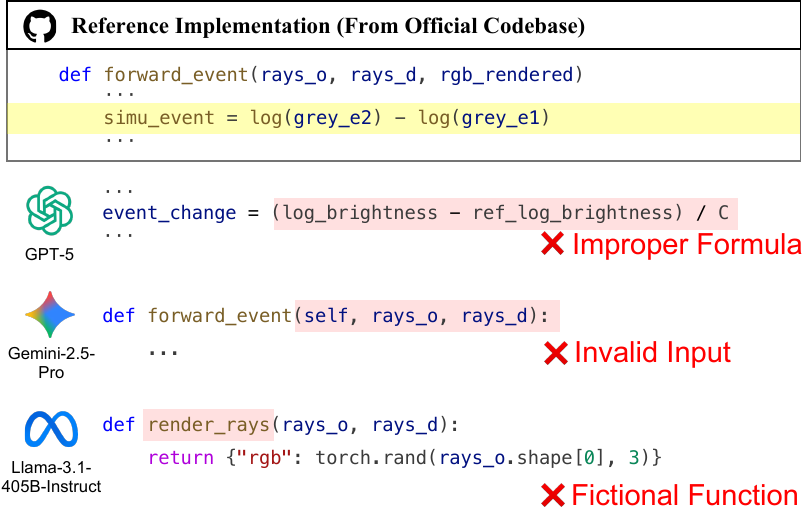}
    \caption{\textbf{Case Study: Consistent Failure Across LLMs on a Simple Function.}
    The function \texttt{forward\_event} approximates “event accumulation” using the logarithmic intensity difference derived from two event-camera frames. Despite its brevity and simplicity, all tested LLMs failed.
    }
    \label{fig:case1}
\end{figure}

\paragraph{Key Observations.}
As shown in Table~\ref{tab:main}, we summarize four key trends:
\textbf{(1)} GPT-5 achieves the highest overall score (36.6\%, showing balanced strength across both \textit{Geometric Transformations} and \textit{Mechanics/Optics Formulation}, reflecting its superior general 3D reasoning ability.
\textbf{(2)} Claude-Sonnet-4.5 follows closely, with consistently strong results in \textit{geometric
logic routing} and competitive mathematical reasoning, indicating robust spatial comprehension but slightly weaker algorithmic fidelity.
\textbf{(3)} In contrast, open-source models such as Qwen3-Coder-480B and DeepSeek-R1 show partial competence in individual categories but underperform in \textit{Algorithm Implementation}, suggesting limitations in multi-step reasoning and code generation.
\textbf{(4)} Across all models, scores on \textit{Algorithm Implementation} and \textit{Geometric Logic Routing} are significantly lower than on geometric or mathematical categories, revealing a persistent gap between theoretical understanding and executable research-level implementation.
This difficulty is clearly exemplified in Figure~\ref{fig:case1}, which shows that every tested model failed to correctly implement even a very simple function, \texttt{forward\_event} from~\cite {huang2025inceventgs}, due to errors ranging from improper formula usage to invalid input parameters and completely fictional code.

\begin{tcolorbox}[
    colback=pink!10,       
    colframe=red!70!black, 
    coltitle=white,        
    fonttitle=\bfseries,   
    title=Key Finding 1,       
    arc=6pt,               
    boxrule=1pt,           
    left=4pt, right=4pt, top=4pt, bottom=4pt, 
]
Best LLM (GPT-5) leads overall (36.6\%) yet achieves only around one-third of the ideal score, revealing a substantial gap between 3D understanding and executable implementation across all models.
\end{tcolorbox}

\begin{figure}[h]
    \centering
    \includegraphics[width=\linewidth]{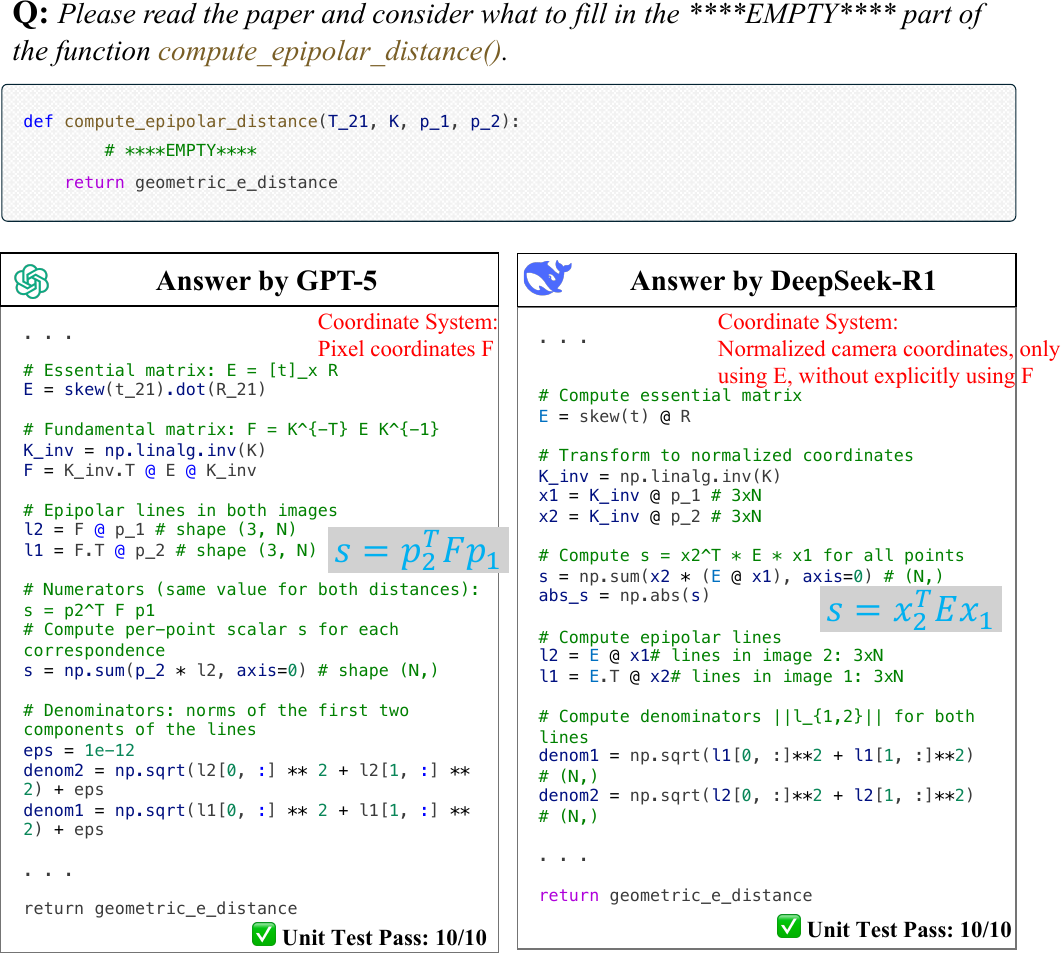}
    \caption{\textbf{Case Study: Creativity Correctness.} The function \texttt{compute\_epipolar\_distance} requires calculating the symmetric epipolar distance between corresponding image points $\mathbf{p}_1$ and $\mathbf{p}_2$ given $T_{21}$ and $\mathbf{K}$.
    GPT-5 uses the Fundamental Matrix ($\mathbf{F}$) method ($\mathbf{l}_2 = \mathbf{F}\mathbf{p}_1$), operating directly on pixel coordinates.
    DeepSeek-R1, conversely, first transforms the inputs to normalized coordinates ($\mathbf{x}_1, \mathbf{x}_2$) and then applies the Essential Matrix ($\mathbf{E}$) method ($\mathbf{l}'_2 = \mathbf{E}\mathbf{x}_1$).
    Both approaches are mathematically equivalent ($\mathbf{F} = \mathbf{K}^{-T} \mathbf{E} \mathbf{K}^{-1}$) and yield the correct final distance, thus demonstrating \textbf{Creative Correctness}: models select distinct, valid pathways to achieve the required $3\text{D}$ geometry constraint.
    }
    \label{fig:case_creative_correctness}
\end{figure}

However, for some cases that LLMs answer correctly, we observe a phenomenon of \textit{Creative Correctness}, where models like GPT-5 and DeepSeek-R1 solve the same problem (\textit{e.g.}, \texttt{compute\_epipolar\_distance}) using mathematically equivalent but distinctly different implementations (using the Fundamental Matrix $\mathbf{F}$ vs. the Essential Matrix $\mathbf{E}$ on normalized coordinates), as detailed in Figure~\ref{fig:case_creative_correctness}.

\begin{tcolorbox}[
    colback=pink!10,       
    colframe=red!70!black, 
    coltitle=white,        
    fonttitle=\bfseries,   
    title=Key Finding 2,       
    arc=6pt,               
    boxrule=1pt,           
    left=4pt, right=4pt, top=4pt, bottom=4pt, 
]
LLMs can achieve Creative Correctness in 3D geometry tasks, successfully implementing distinct but mathematically equivalent code paths for the same problem.
\end{tcolorbox}

\subsection{General vs. Research Capability}
\label{subsec:ability}

\begin{figure}[h]
    \centering
    \includegraphics[width=1.0\linewidth]{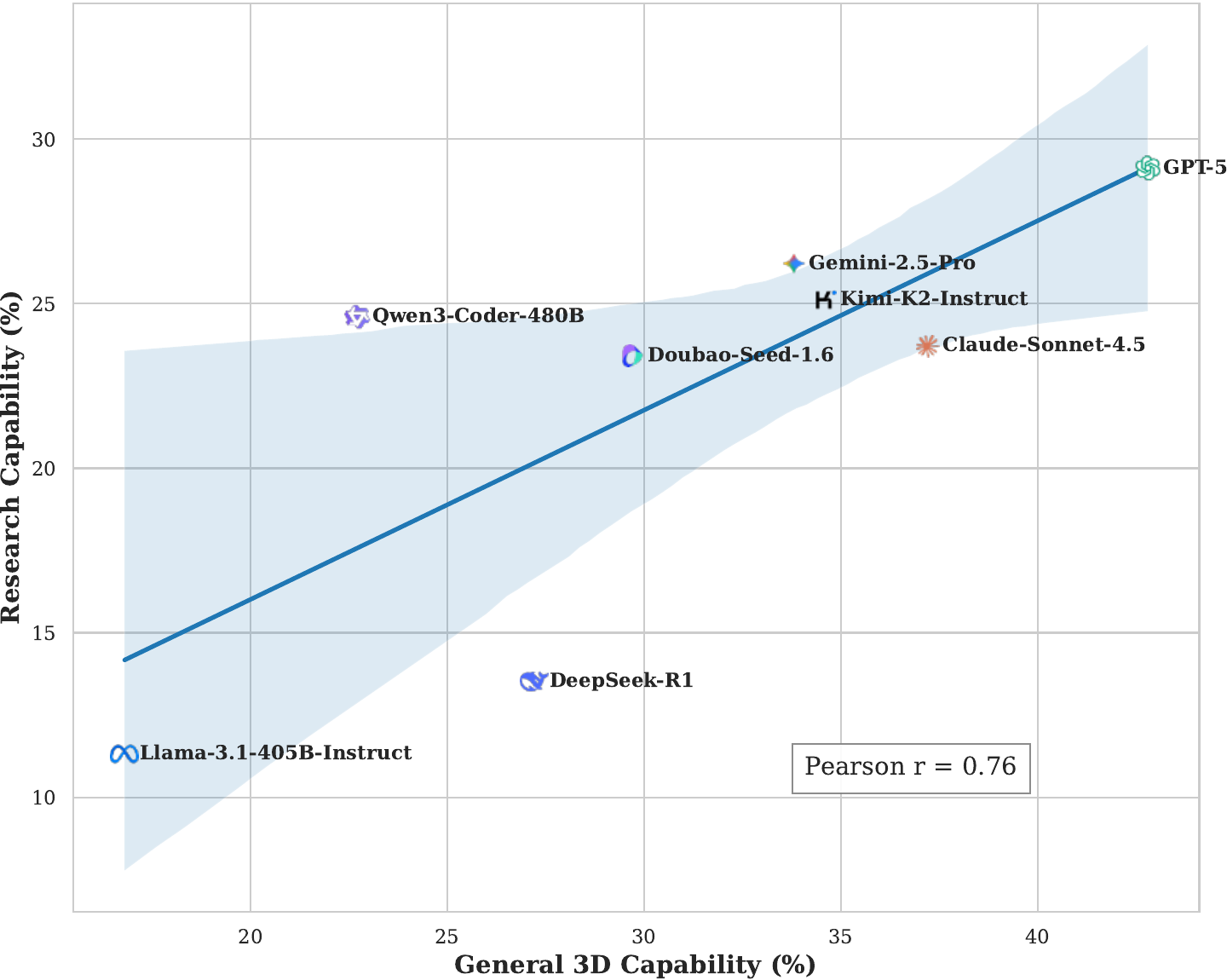}
    \caption{Comparison of General and Research Capability.
    }
    \label{fig:general-vs-research}
\end{figure}
\begin{figure}[h]
    \centering
    \includegraphics[width=\linewidth]{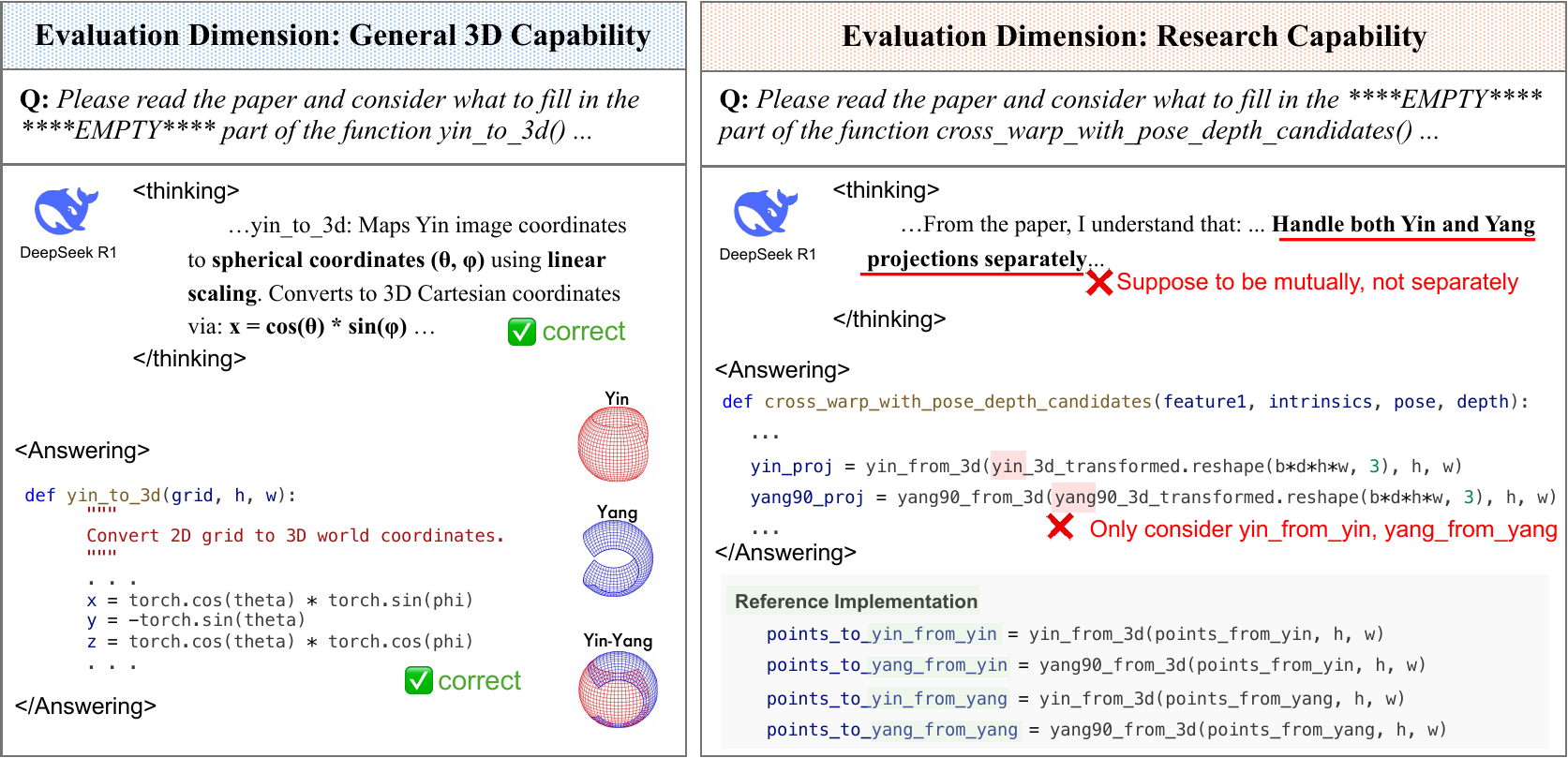}
    \caption{
        \textbf{Case Study: Divergence between General and Research Capability.} 
        Left: DeepSeek-R1 correctly solves the basic geometric problem of mapping 2D image coordinates to 3D spherical coordinates using trigonometric functions, demonstrating strong foundational knowledge. 
        Right: In contrast, the model fails to correctly implement the paper-specific function \texttt{cross\_warp\_with\_pose\_depth\_candidates}. The error involves misinterpreting the required mutual projection logic (Yin $\leftrightarrow$ Yang) as separate projections, highlighting the difficulty LLMs face in procedural reasoning and fine-grained algorithmic synthesis based on paper content.
    }
    \label{fig:case2}
\end{figure}

\paragraph{Key Observations.}
As illustrated in Fig.~\ref{fig:general-vs-research}, the results exhibit a clear gap between \textit{general 3D capability} and \textit{paper-specific research capability}, with a \textbf{positive correlation} (\textit{Pearson} $r=0.76$) between the two dimensions.
\textbf{(1)} Across all models, general 3D understanding is always higher than research-oriented reproduction, suggesting that foundational geometric and mathematical knowledge is more accessible to current LLMs than the procedural reasoning required for algorithmic synthesis.
As shown in the case study with DeepSeek-R1 (Fig.~\ref{fig:case2}), DeepSeek-R1 successfully handles basic, general 3D problems but fails on paper-specific research problems.
\textbf{(2)} Some models, e.g., Claude-Sonnet-4.5, dominate in general capability (37.2\%), yet their research-specific performance lags behind (23.7\%), indicating difficulty in transferring theoretical understanding into implementation.
\textbf{(3)} In contrast, Qwen3-Coder-480B show more balanced profiles (22.7/24.6\%), implying stronger code-level compositional reasoning despite weaker foundational abstraction.
\textbf{(4)} Open-source models such as DeepSeek-R1 and Llama-3.1-405B-Instruct exhibit the steepest decline, reflecting the absence of sufficient research-context training signals.
Notably, even the best models achieve less than 45\% on either axis, underscoring the overall immaturity of current LLMs in scientific 3D reasoning.

\paragraph{Insights.}
This breakdown reveals that most LLMs possess relatively solid 3D priors and mathematical intuition but struggle to internalize algorithmic logic from research contexts.  
\textbf{(1)} The benchmark’s two-axis design effectively isolates this cognitive bottleneck, quantifying the translation gap from geometric reasoning to executable research workflows.  
\textbf{(2)} Fundamentally, this gap stems from limited compositional generalization and weak procedural abstraction in current LLM architectures, suggesting that bridging symbolic geometry and structured program synthesis remains an open challenge for “scientific” language models.

\begin{tcolorbox}[
    colback=pink!10,       
    colframe=red!70!black, 
    coltitle=white,        
    fonttitle=\bfseries,   
    title=Key Finding 3,       
    arc=6pt,               
    boxrule=1pt,           
    left=4pt, right=4pt, top=4pt, bottom=4pt, 
]
LLMs show relatively stronger general 3D understanding than research-level capability, and a clear gap between knowing and implementing 3D geometry.
\end{tcolorbox}

\begin{figure}[htbp]
    \centering
    \includegraphics[width=\linewidth]{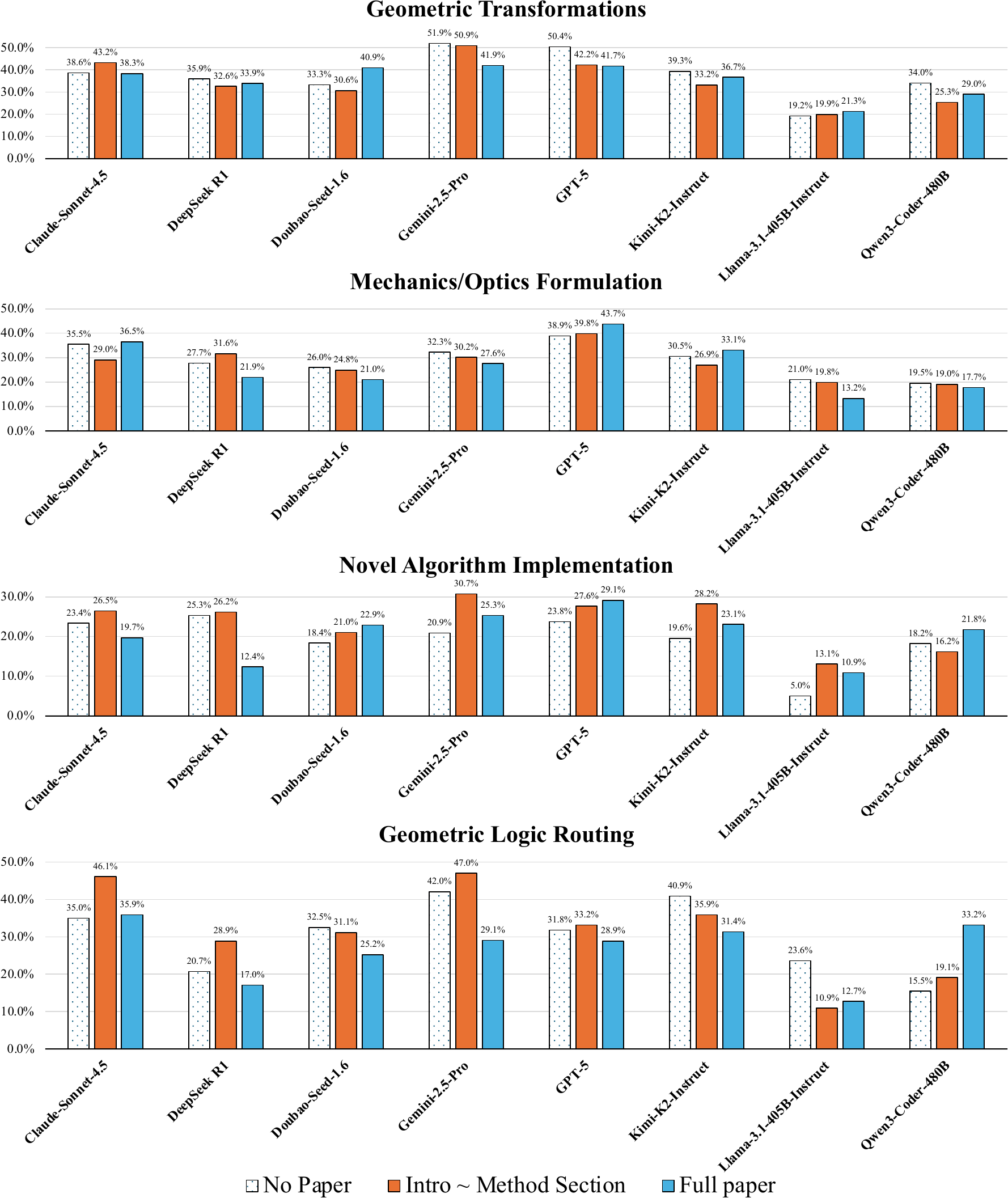}
    \caption{Impact of Paper Length on Detailed Dimension Performance.
    }
    \label{fig:paper-length-detail}
\end{figure}

\subsection{Impact of Paper Length}
\label{subsec:paper-sensitivity}

While research papers provide essential methodological cues, they also introduce significant textual noise—long introductions, experimental details, and unrelated context may overwhelm an LLM’s limited attention span.
To examine whether more information truly helps, we measure each model’s performance under three paper input conditions:
\textit{full paper}, \textit{cutting off at the Method section}, and \textit{no paper}.
This setting allows us to analyze how effectively LLMs utilize long-context research information, both at the overall level and within each capability dimension.

\begin{figure}[h]
    \centering
    \includegraphics[width=\linewidth]{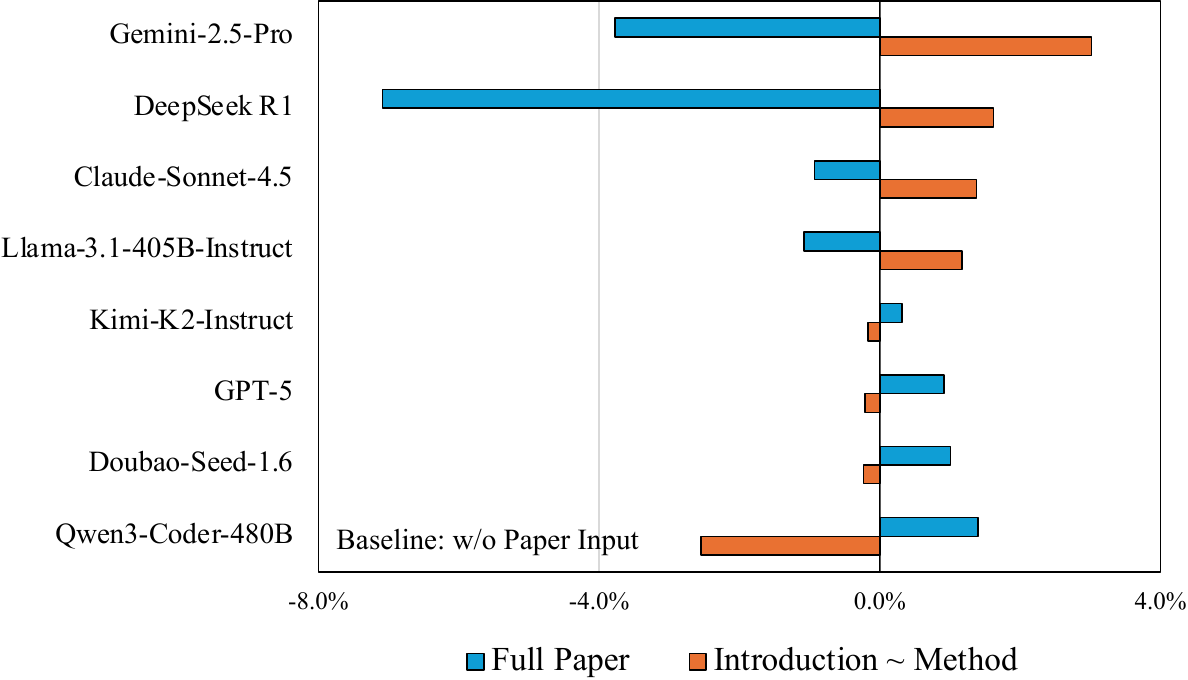}
    \caption{Impact of Paper Length on Overall Performance.
    }
    \label{fig:paper-length-overall}
\end{figure}


Overall, the results reveal that \textit{more information does not necessarily lead to better reasoning.}
When the full paper is provided, most models show stagnant or even decreased performance compared to the up-to-method setting, suggesting that long text often introduces redundant or distracting signals.
Notably, Gemini-2.5-Pro and DeepSeek-R1 achieve their peak scores when restricted to method sections, implying that these models can identify algorithmic cues once non-essential context is removed.
By contrast, models such as Qwen3-Coder-480B and Doubao-Seed-1.6 benefit more from additional text, as providing more context improves final performance.
GPT-5 and Kimi-K2-Instruct remain comparatively robust, showing stable results across all settings, suggesting that their comprehension relies less on text quantity and more on internalized geometric priors.



A deeper breakdown by category further reveals a divergence between \textit{general} and \textit{research} skills.
Tasks under \textit{Geometric Transformation} and \textit{Mechanics/Optics Formulation}—which test universal 3D knowledge—show little or no improvement when more paper content is added, implying that these abilities are largely intrinsic and independent of contextual input.
In contrast, \textit{Algorithm Implementation} and \textit{Geometric Logic Routing}—which depend on understanding paper-specific logic—are highly sensitive to paper length: 
moderate-length (up-to-method) inputs yield the best results, whereas full papers often lead to a performance drop due to narrative noise and inconsistent formatting.
This pattern confirms that excessive textual context can hinder procedural reasoning instead of helping it.

\begin{tcolorbox}[
    colback=pink!10,       
    colframe=red!70!black, 
    coltitle=white,        
    fonttitle=\bfseries,   
    title=Key Finding 4,       
    arc=6pt,               
    boxrule=1pt,           
    left=4pt, right=4pt, top=4pt, bottom=4pt, 
]
Providing more text does not guarantee better understanding—most LLMs perform best with concise, method-focused inputs, revealing limited long-context comprehension.
\end{tcolorbox}

\subsection{Failure Case Analysis}

\begin{figure}[h]
    \centering
    \includegraphics[width=\linewidth]{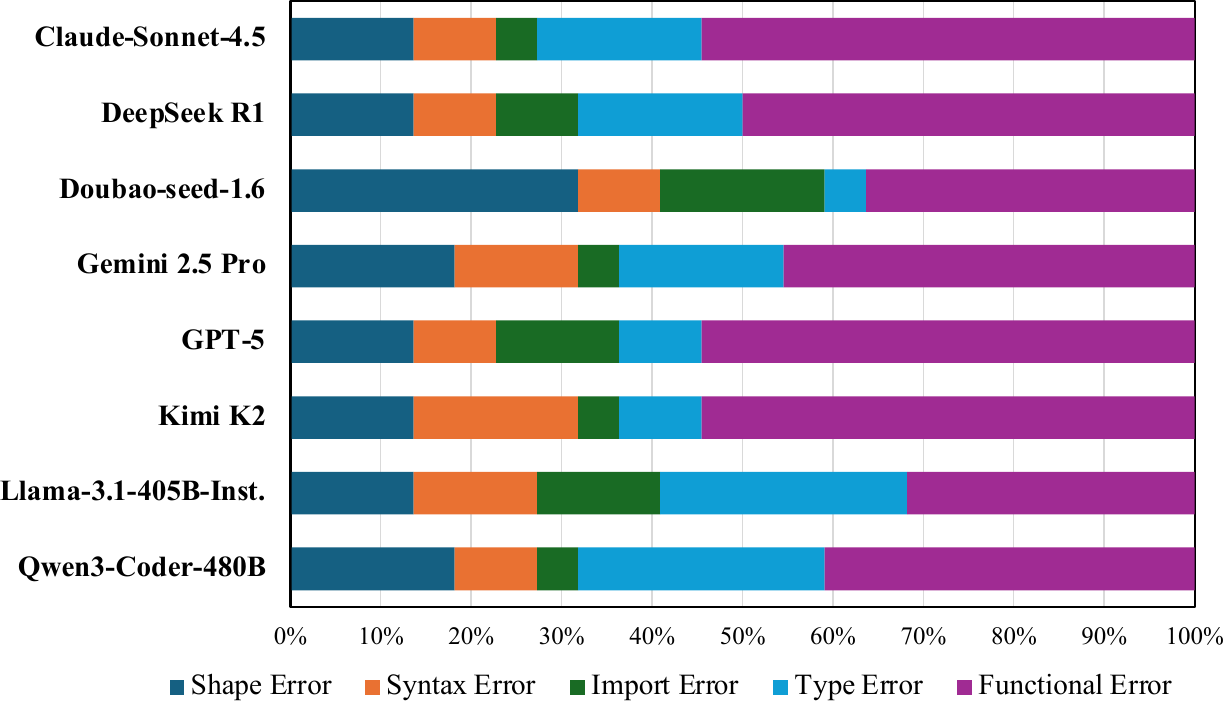}
    \caption{Distribution of Common Failure Types across Models on GeoCodeBench.}
    \label{fig:failure-type}
\end{figure}

To better understand model weaknesses on GeoCodeBench, we categorize all incorrect generations into five major failure types, as shown in Figure~\ref{fig:failure-type}.
Across all models, \textit{functional errors} dominate, showing that even when the code compiles, the algorithmic logic often fails to reproduce the intended behavior.
\textit{Type} and \textit{shape errors} are also frequent, revealing insufficient understanding of tensor dimensions and geometry-aware data structures that are fundamental to 3D vision.
In contrast, \textit{syntax} and \textit{import errors} occur less often but are more prevalent in smaller or less instruction-tuned models, indicating weaker robustness in code organization and dependency handling.
Overall, even leading models struggle with multi-step geometric reasoning and functional consistency, underscoring the intrinsic difficulty of 3D vision code generation in GeoCodeBench.

\section{Conclusion}
GeoCodeBench establishes the first rigorous, execution-based benchmark for evaluating whether LLMs can implement real 3D geometric vision code from scientific text. Built from expert-selected functions sourced from recent top-venue 3D vision papers and paired with high-variability, geometry-aware unit tests, the benchmark captures both fundamental operators and research-level algorithmic logic. Our results reveal a substantial capability gap: even the strongest models show limited reliability, struggle with long-context comprehension, and falter on higher-level reasoning. At the same time, instances of creative yet correct implementations demonstrate meaningful space for future progress. We hope GeoCodeBench provides a solid foundation for developing LLMs that can truly understand, reason about, and implement complex 3D vision algorithms.

{
    \small
    \bibliographystyle{ieeenat_fullname}
    \bibliography{main}

}

\clearpage
\setcounter{page}{1}
\maketitlesupplementary


\section{More Statistics of GeoCodeBench}
To provide a clearer understanding of the composition and characteristics of GeoCodeBench, we first summarize the data sources and scale of the benchmark. As shown in Table~\ref{tab:repo_stat}, GeoCodeBench collects 47 repositories from CVPR'25, ICCV'25, and ICLR'25, resulting in a total of 100 problem instances, with CVPR'25 contributing the largest portion. 

\begin{table}[h]
\centering
\caption{Data Source Statistic of GeoCodeBench.}
\label{tab:repo_stat}
\small
\begin{tabularx}{0.8\linewidth}{Xcc}
\toprule
\textbf{Source} & \textbf{Repos Nums} & \textbf{Problems Nums} \\ \midrule
CVPR'25 & 28 & 55 \\
ICCV'25 & 15 & 33 \\
ICLR'25 & 4  & 12 \\ \midrule
Total & 47  & 100 \\ \bottomrule

\end{tabularx}
\end{table}

Furthermore, we analyze the token statistics of different components within the benchmark, including the structured paper content, code-with-masked functions, and golden implementation functions (Table~\ref{tab:token_stats}). The structured paper content is significantly longer than other components, with an average length of 20{,}232 tokens, whereas the golden implementation functions remain concise, averaging only 753 tokens. 

\begin{table}[h]
\centering
\caption{Tokens Statistic of Different Components in GeoCodeBench.}
\label{tab:token_stats}
\small
\begin{tabular}{lccc}
\toprule
\textbf{Component} & \textbf{Mean} & \textbf{Min} & \textbf{Max} \\
\midrule
Structured Paper Content & 20,232 & 14,052 & 31,079 \\
Code w/ Masked Function & 3,802 & 68 & 14,036 \\
Golden Implementation Function & 753 & 222 & 2,917 \\
\bottomrule
\end{tabular}
\label{tab:filesize}
\end{table}

\begin{figure}[h]
    \centering
    \includegraphics[width=1.0\linewidth]{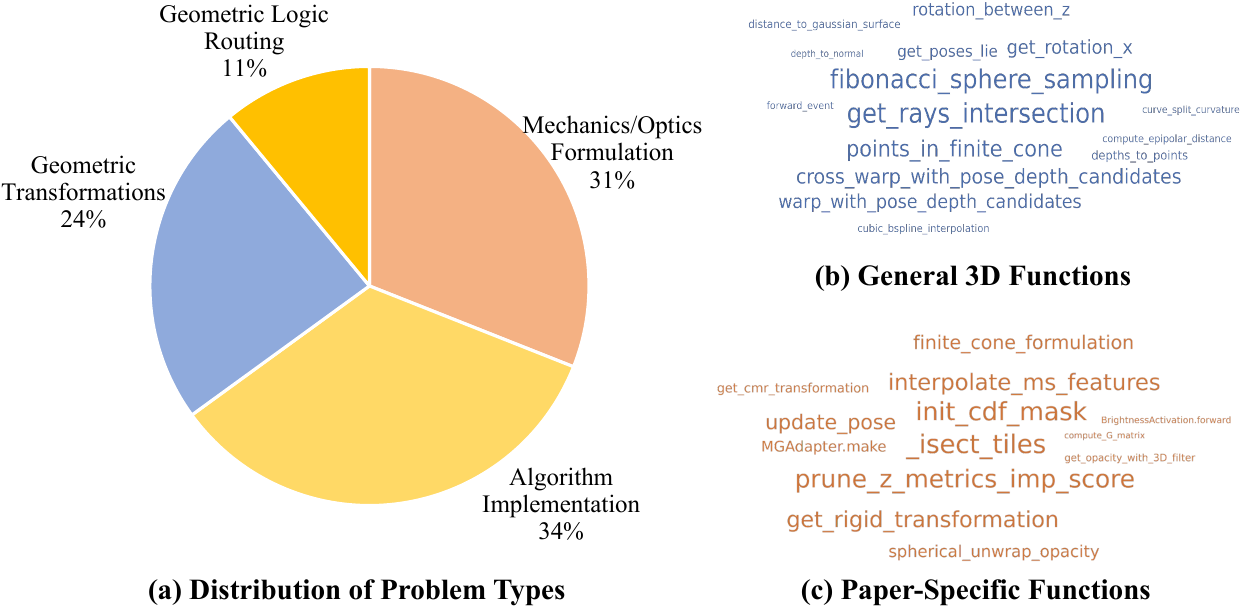}
    \caption{Problem Type and Representative Functions in GeoCodeBench.}
    \label{fig:data_statistic}
\end{figure}

GeoCodeBench also features diverse problem types, as illustrated in Fig.~\ref{fig:data_statistic}(a). Novel algorithm implementations tasks constitute the largest proportion (34\%), followed by Mechanics/Optics formulation(31 \%),  geometric transformations (24\%), and Geometric Logic Routing tasks (11\%). Representative keywords for general 3D functions and paper-specific research ability functions are visualized in Fig.~\ref{fig:data_statistic}(b)--(c), demonstrating the benchmark's coverage of essential 3D geometry operations, such as spatial transformations, geometric intersections, optimization procedures, and problem-specific functional modules. Collectively, these statistics indicate that GeoCodeBench offers comprehensive diversity in task sources, content structure, and functional requirements, providing a robust foundation for evaluating PhD-level coding capabilities in 3D geometric computer vision.

\section{Details of Evaluation}

\begin{figure*}[t]
\begin{lstlisting}[
    breaklines=true,
    breakatwhitespace=true,
    numbers=left,
    numberstyle=\tiny,
    frame=single,
    basicstyle=\footnotesize\ttfamily
]
You are an expert in 3D vision. Please read the paper and think about what to fill in the ****EMPTY**** section of the code. 
Output format: Returns the function according to the template.

## PAPER:
{paper_content_json}

## CODE with EMPTY:
{code}

## TEMPLATE:
{template}

Please fill out the template above.
Your response must strictly follow the format below:

<thinking>
[Your reasoning or thinking content here]
</thinking>

<answering>
[Fill in the complete template here, ensuring that:
 1.All required imports and components are included,
 2.The structure strictly follows the provided template,
 3.The result is a complete, ready-to-use implementation]
</answering>

\end{lstlisting}
\captionof{lstlisting}{Input Prompt for LLM.}
\label{lst:prompt}
\end{figure*}

For each problem in GeoCodeBench, we first construct the full prompt following the format shown in Listing~\ref{lst:prompt}. The prompt contains (1)~the structured paper content, (2)~the masked implementation function that the model must rewrite, and (3)~the template that defines the I/O and necessary packages. 

For every problem instance, we invoke the target LLM exactly once using greedy decoding. The model's response is parsed and inserted into the corresponding code template to reconstruct the full implementation. We then execute the unit tests associated with that problem to determine whether the generated solution is functionally correct. Each problem contributes a unit test PassRate score.
After evaluating all problems, we compute the final score for each model by averaging the PassRate values across the entire benchmark. 

To ensure fairness across models, we standardize several important evaluation conditions: (1)~all models are evaluated with identical prompts and test suites; (2)~no system prompts or model-specific engineering tricks are used; (3)~for unit test results, we set tolerance and precision thresholds to account for minor numerical variations; (4)~ for some problems, the reference implementations are not the only correct implementations which derived directly from the original templates. For these problems, we either impose additional constraints on the templates or relax the unit test acceptance criteria.


\section{More Case Study}

\begin{figure}[h]
    \centering
    \includegraphics[width=\linewidth]{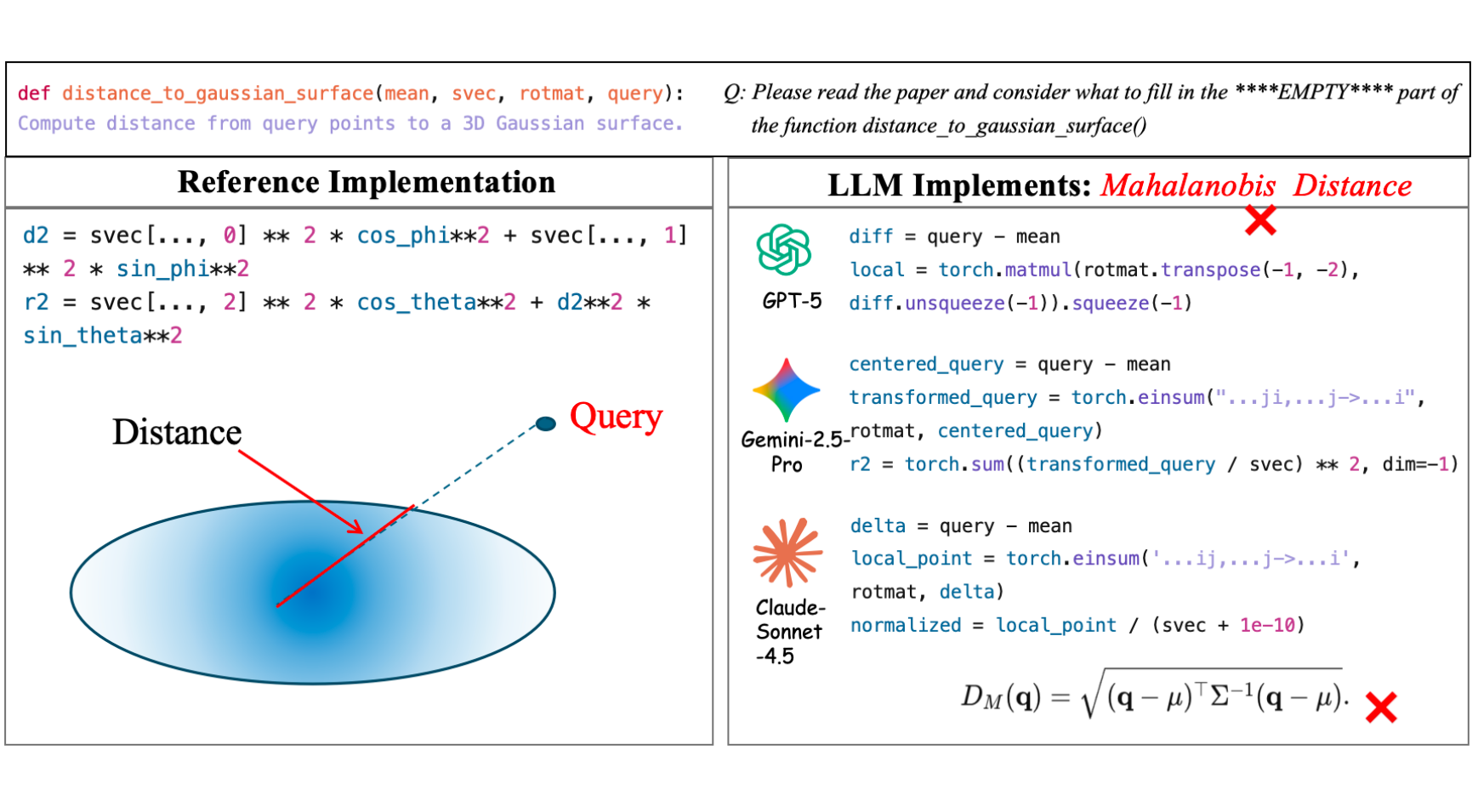}
    \caption{Case Study: Incorrect Type of Distance Implemented. }
    
    \label{fig:paper-improve}
\end{figure}

\begin{figure}[h]
    \centering
    \includegraphics[width=\linewidth]{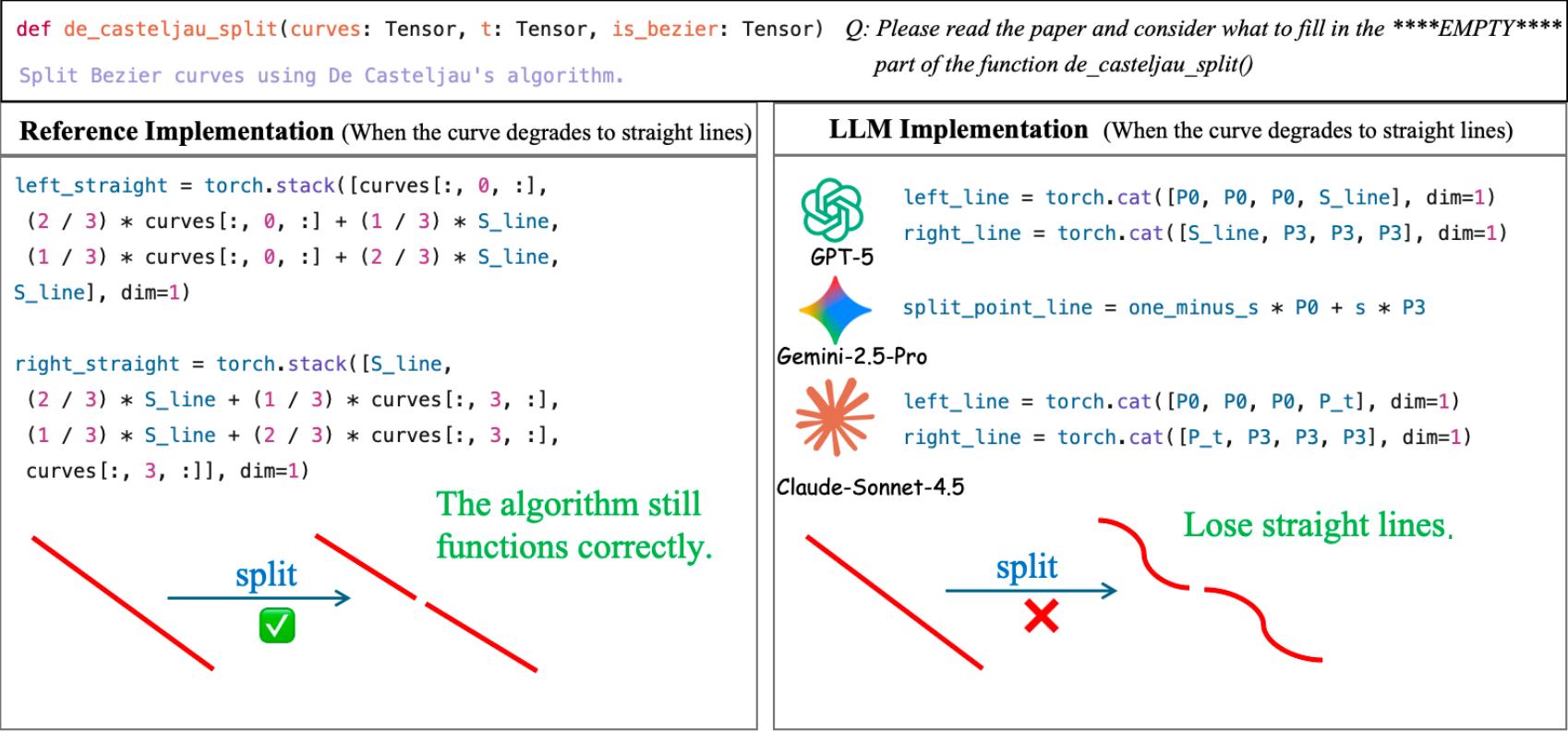}
    \caption{Case Study: Errors of De Casteljau Split When Degrading to Lines. 
    }
    \label{fig:case5}
\end{figure}

\begin{figure}[h]
    \centering
    \includegraphics[width=\linewidth]{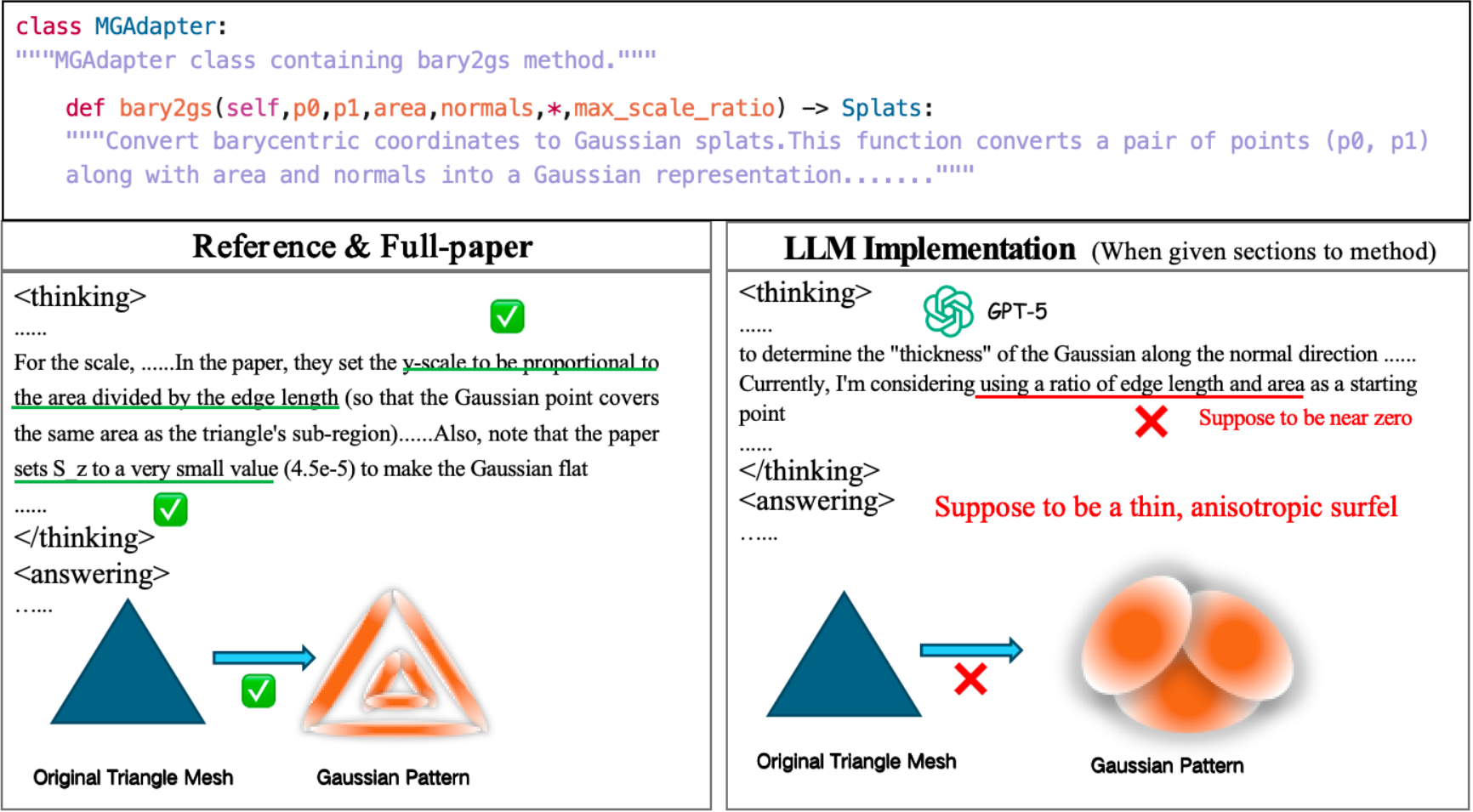}
    \caption{Case Study: MGadapter Fidelity Differs with Text Length.}
    
    \label{fig:paper-sensitivity1}
\end{figure}

\begin{figure}[h]
    \centering
    \includegraphics[width=\linewidth]{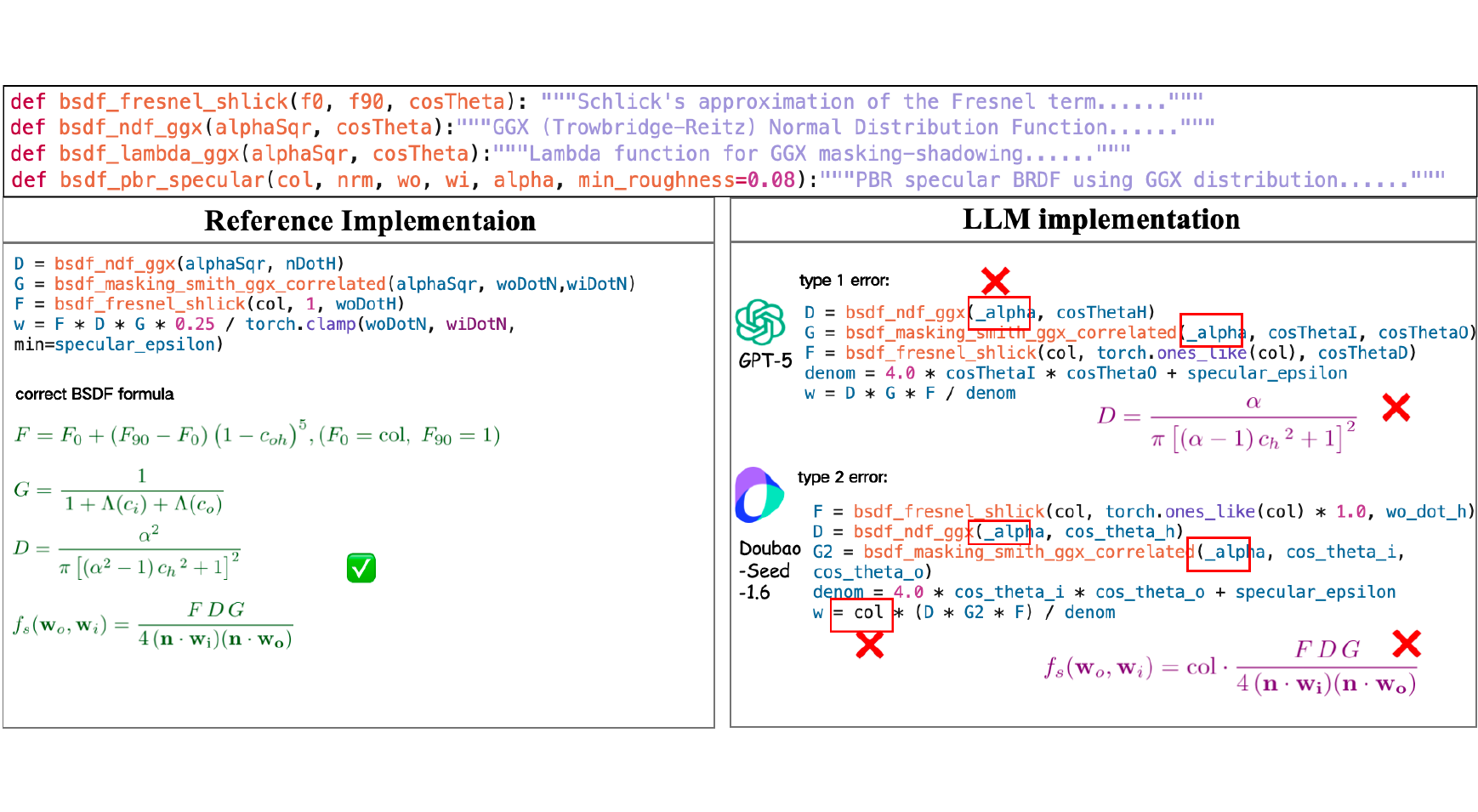}
    \caption{Case Study: Incorrect Implementation of the BRDF Formulas.
    }
    \label{fig:paper-sensitivity2}
\end{figure}

\begin{figure}[h]
    \centering
    \includegraphics[width=\linewidth]{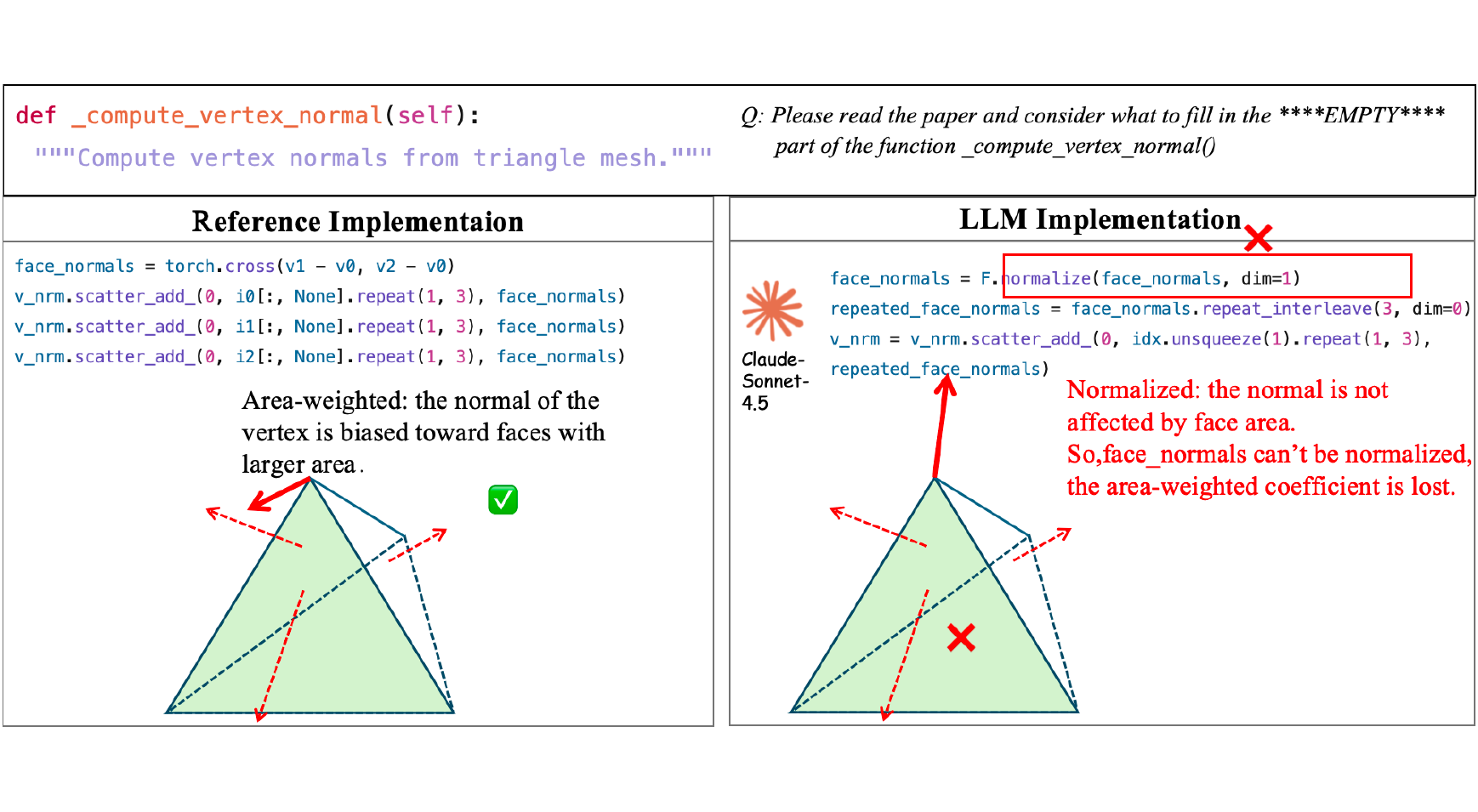}
    \caption{Case Study: Incorrect Normalization that Ignores Area-weighting Coefficients.
    }
    \label{fig:paper-sensitivity}
\end{figure}

\paragraph{Case Study: Misinterpreting “Distance to a 3D Gaussian Surface” as Mahalanobis Radius}

This case comes from\cite{shen2025dof}, which evaluates the ability of LLMs to implement a ``distance to a 3D Gaussian surface'' function in a three-dimensional geometry and linear algebra setting.

In the reference implementation, the function \texttt{distance\_to\_gaussian\_surface} is defined as follows: it returns the distance from the Gaussian center along the ray $\text{query} - \text{mean}$ to the 1-sigma ellipsoidal surface, i.e., the surface radius of the ellipsoid in that direction. This quantity is independent of the actual radial position of the query point and represents a true physical distance that can be directly used for geometric decisions. 

By contrast, the LLM-generated implementation computes the Mahalanobis radius from the point to the center of the ellipsoid, which is a normalized, dimensionless distance. This value must be further transformed to fit the logic of the current geometric pipeline. It is more naturally suited to statistical analysis and anomaly detection, rather than to direct geometric distance computation.

From a code quality perspective, the LLM implementation correctly handles translation by the mean, rotation into the local frame, scaling by the axis lengths, and broadcasting across batch dimensions. Mathematically, it is internally consistent and implements a standard and widely used notion of ``distance to a Gaussian''. However, from the standpoint of the benchmark, it exhibits a critical geometric semantic error: the distance measure it computes does not match the definition used in the reference implementation. As a result, on our designed test cases, the numerical outputs of the two implementations diverge systematically, and the LLM solution is judged incorrect.

\paragraph{Insight}
This case exposes a typical failure mode of LLMs in 3D geometry--oriented code generation: while they are strong at implementing canonical mathematical constructs, they rely heavily on natural-language specifications and struggle to infer task-specific implicit semantics. Consequently, they tend to default to the most common textbook definition---here, the Mahalanobis distance---rather than the problem-specific ``directional surface radius'' actually required by the dataset and evaluation. See figure \ref{fig:paper-improve}.

\begin{tcolorbox}[
    colback=pink!10,       
    colframe=red!70!black, 
    coltitle=white,        
    fonttitle=\bfseries,   
    title=Key Finding 5,       
    arc=6pt,               
    boxrule=1pt,           
    left=4pt, right=4pt, top=4pt, bottom=4pt, 
]
LLMs tend to default to the most common textbook definition rather than the problem-specific implementation actually required by the paper, which reflects the limitation of reasoning ability of a statistical model.
\end{tcolorbox}

\paragraph{Case Study: Preserving Straight Lines in De Casteljau-Based Bezier Splitting}

This case comes from \cite{gao2025curve}, designed to evaluate the ability of LLMs to implement the Bezier curve splitting function \texttt{de\_casteljau\_split} in a three-dimensional geometric setting. The function takes as input a batch of curve control points $\text{curves} \in \mathbb{R}^{B \times D \times 4}$, a split parameter $t$,  a Boolean mask, and it outputs the left and right sub-curves, each represented in a unified four-control-point form. In the benchmark, the reference implementation and the LLM-generated implementations are essentially equivalent in their De Casteljau recursion for true Bezier curves; however, they exhibit a critical geometric semantic discrepancy in the representation convention for straight-line segments (non-Bezier cases).

In the reference implementation, curves are split using the standard cubic Bezier De Casteljau algorithm, yielding left and right segments. For straight-line segments, the resulting sub-curves are still geometrically straight: all four control points are strictly collinear, and the parameterization remains consistent with the canonical representation of a straight line as a cubic Bezier curve. This implicit constraint is crucial for subsequent geometric processing.

In contrast, the LLM implementation correctly applies the De Casteljau recursion in the Bezier case, but adopts a more ``intuitive yet pipeline-incompatible'' strategy for straight lines: the two intermediate control points are simply duplicated from the endpoints, violating the system's representational convention. When interpreted as cubic Bezier curves, such control-point configurations cause the curve to ``collapse'' near the endpoints and deviate from a true straight line.

\paragraph{Insight}
This case shows that the LLM is capable of reproducing the local De Casteljau splitting pattern, but struggles with preserving global representation invariants and downstream contracts. It defaults to a locally simple control-point layout that is mathematically valid yet semantically incompatible with the canonical line-encoding convention, demonstrating that the model’s code generation is driven by local plausibility rather than by a consistent understanding of the project’s geometric conventions. See figure \ref{fig:case5}.

\paragraph{Case Study: Context Length and Geometry–Gaussian Consistency}

This task evaluates whether LLMs can implement the MGadapter-style mapping from mesh-sampled geometric elements to 3D Gaussian splats, following the design used in GeoSplatting\cite{GeoGaussian}. The target function \texttt{bary2gs} takes two points on a triangle edge ($p_0$, $p_1$), a per-sample area, and a surface normal, and returns a \texttt{Splats} object (means, anisotropic scales, orientation quaternions, colors, and opacities). The correct implementation encodes a very specific geometric intent: each splat should behave as a thin surfel aligned with the local mesh geometry---its long axis aligned with the edge direction in the tangent plane, its short in-plane axis derived from the projected area, and an almost-zero thickness along the normal (log scale fixed at $-10$) to enforce a consistent geometry--Gaussian correspondence.

When given the \textbf{full paper}, the LLM implementation qualitatively captures much of this intent. It correctly places the means at the edge midpoint, constructs an orthonormal frame from a projected edge direction and the normal, and assigns anisotropic scales where the major axis depends on edge length and the minor axis depends on area, with a very small thickness along the normal. 

However, in the \textbf{to-method} setting, the implementation drifts further toward a generic 3DGS heuristic. The LLM still constructs a local frame from the normal and the edge direction and uses edge length and area to define three scales, but the coupling between these quantities and the axes is looser: some implementations tend to isotropic Gaussian and some set a non-zero value for thickness. The resulting splats are less tightly constrained by the triangle area---making them less faithful to the MGadapter's core design.

\paragraph{Insight}
This case highlights two effects of context length on LLM code generation. With richer context, the model more accurately reconstructs the intended anisotropic surfel parameterization and uses the right geometric ingredients. With reduced context, it reverts to a more generic ellipsoid construction that is mathematically sound but no longer matches the specialized geometry--Gaussian consistency of the original MGadapter. See figure \ref{fig:paper-sensitivity1}.

\paragraph{Case Study: GGX Roughness Semantics Mismatch}

This case comes from \cite{zhu2025gaussian}, designed to evaluate the ability of LLMs to implement a physically correct specular BRDF based on the GGX (Trowbridge--Reitz) microfacet model. The reference implementation provides a complete set of BSDF components, including Schlick Fresnel (\texttt{bsdf\_fresnel\_shlick}), the GGX normal distribution function (\texttt{bsdf\_ndf\_ggx}), the GGX masking--shadowing lambda term (\texttt{bsdf\_lambda\_ggx}), Smith's correlated masking function (\texttt{bsdf\_masking\_smith\_ggx\_correlated}), and the final PBR specular BRDF \texttt{bsdf\_pbr\_specular}. The interface contract is explicit: the first argument of these GGX helper functions is $\alpha^2$ (the square of the roughness parameter), while the externally exposed \texttt{alpha} is squared to obtain \texttt{alphaSqr}, which is propagated throughout the GGX computation chain.

At first glance, the LLM-generated implementation appears very similar to the reference code, but it mishandles the semantics of the roughness parameter. The LLM implementation retains the name \texttt{alphaSqr}, but passes \texttt{\_alpha} directly from \texttt{bsdf\_pbr\_specular} without squaring it. As a consequence, the entire GGX computation chain is numerically equivalent to using $\alpha$ in place of $\alpha^2$, thereby altering the mapping between the user-visible roughness parameter and the underlying GGX slope distribution. This leads to systematically different specular lobe shapes compared to the reference, especially in the low-roughness regime. Although each function looks locally correct in terms of its formulae, the implementation deviates from the ground-truth behavior from the perspective of the API contract.

\paragraph{Insight}
This case highlights a broader limitation of current LLMs in scientific and graphics code: they are very good at reconstructing familiar mathematical structures and formulations, but much less reliable at tracking subtle parameter semantics and system-level contracts that are changeable in the codebase. In particular, they tend to preserve the “shape” of an algorithm while drifting on the meaning of its parameters, revealing a gap between surface-level code synthesis and a deeper understanding of physical models and engineering interfaces. See figure \ref{fig:paper-sensitivity2}.

\paragraph{Case Study: Loss of Area Weighting in Vertex Normal Computation}

This case comes from \cite{guo2024tetsphere}, designed to evaluate whether LLMs, when computing vertex normals on triangle meshes, can simultaneously reproduce both the geometric details and the engineering robustness of the reference implementation. The reference implementation first uses the triangle index buffer \texttt{t\_pos\_idx} to fetch the three vertex positions \texttt{v0}, \texttt{v1}, and \texttt{v2} for each face, and then computes the face normals \texttt{face\_normals} via \texttt{torch.cross(v1 - v0, v2 - v0)}. These face normals are \emph{not} normalized; their magnitudes are proportional to the corresponding triangle areas. Consequently, when these face normals are splatted to vertices and accumulated, the method naturally implements an \emph{area-weighted} vertex normal estimation: large triangles contribute more to their incident vertex normals, while small triangles contribute less. Before normalization, the reference implementation explicitly handles degenerate cases: for vertices whose accumulated normal has (almost) zero norm, the normal is replaced by a default direction \texttt{[0, 0, 1]}, and only then passed through \texttt{F.normalize}. In this way, even in the presence of degenerate triangles or ``isolated'' vertices (vertices not referenced by any face), the output is guaranteed to consist of finite unit normals, satisfying the assumptions made by subsequent rendering and geometry processing stages.

The LLM-generated implementation is structurally close to the reference: it also computes face normals from triangle vertices, accumulates them at vertices, and finally applies \texttt{F.normalize} to obtain unit-length normals. However, it deviates from the reference design in two critical aspects. First, the LLM version immediately normalizes the face normals via \texttt{face\_normals = F.normalize(face\_normals, dim=1)}. This forces each triangle to contribute unit-length normals to its incident vertices, thereby discarding the area weighting and making every face contribute equally. On non-uniform meshes (e.g., regions with many small triangles adjacent to fewer large ones), this changes both the smoothing behaviour and the geometric meaning of the vertex normals, and no longer matches the behaviour of the original rendering system. Second, the LLM version does not implement any explicit handling of degeneracies: for vertices whose accumulated normal is zero, \texttt{F.normalize} relies solely on an internal \texttt{eps} for numerical protection, but does not replace those normals with a prescribed default direction. As a result, the output may still contain zero or near-zero normals, causing the unit tests for this task to fail.

\paragraph{Insight}
This case clearly illustrates the type of error that the benchmark seeks to capture. The LLM has a solid grasp of the core geometric routine ``how to compute vertex normals from a triangle mesh'' and can produce code that is structurally correct, shape-compatible, and behaves reasonably on typical inputs. However, it fails to recognize two crucial engineering semantics embodied in the reference implementation: (1) area-weighted smoothing achieved by preserving the magnitudes of face normals, and (2) the explicit injection of a default normal direction for degenerate or isolated vertices to ensure robustness. This gap between formula-level correctness and engineering semantics is precisely what matters in 3D geometry--related CV functions, and it is exactly the capability deficit that this benchmark aims to systematically evaluate. See figure \ref{fig:paper-sensitivity} for the first type of error.

\begin{tcolorbox}[
    colback=pink!10,       
    colframe=red!70!black, 
    coltitle=white,        
    fonttitle=\bfseries,   
    title=Key Finding 6,       
    arc=6pt,               
    boxrule=1pt,           
    left=4pt, right=4pt, top=4pt, bottom=4pt, 
]
LLMs can reliably implement the main body of algorithms, but  under-handle boundary conditions and corner cases, leading to implementations that appear correct yet fail to pass the full unit-test suite.
\end{tcolorbox}


\end{document}